\documentclass[10pt,twocolumn,letterpaper]{article}

\usepackage{cvpr}              % To produce the CAMERA-READY version
\usepackage{graphicx}
\usepackage{amsmath}
\usepackage{amsfonts}
\usepackage{xcolor}
\usepackage[belowskip=-10pt,aboveskip=-1pt]{caption}
\usepackage{mathtools}
\usepackage{enumitem}
\usepackage{bm}
\usepackage{soul}
\usepackage{svg}

\usepackage{algorithm}
\usepackage{algpseudocode}

\usepackage{isotope}
\usepackage{siunitx}  % for units of measure and data in tables
\usepackage{adjustbox}
\usepackage{footnote}
\makesavenoteenv{tabular}
\makesavenoteenv{table}
\usepackage{booktabs, makecell, multirow, threeparttable}

\usepackage[pagebackref,breaklinks,colorlinks]{hyperref}

\usepackage[capitalize]{cleveref}
\crefname{section}{Sec.}{Secs.}
\Crefname{section}{Section}{Sections}
\Crefname{table}{Table}{Tables}
\crefname{table}{Tab.}{Tabs.}

\def\cvprPaperID{6396} % *** Enter the CVPR Paper ID here
\def\confName{CVPR}
\def\confYear{2023}

\newcommand*\rot{\rotatebox{90}}

\def\eg{\emph{e.g.}}
\def\etal{{\em et al.}}
 
\newcommand{\para}[1]{\vspace{.05in}\noindent\textbf{#1}}

\begin{document}

%\title{Histopathology Image Analysis with Heterogeneous Graph Representation Learning}
\title{Histopathology Whole Slide Image Analysis with Heterogeneous Graph Representation Learning}

\author{Tsai Hor Chan$^{1,*}$, Fernando Julio Cendra$^{1,2*}$, Lan Ma$^2$, Guosheng Yin$^{1,3}$, Lequan Yu$^{1}$\\
$^1$Department of Statistics and Actuarial Science, 
The University of Hong Kong\\
$^2$TCL Corporate Research Hong Kong\\
$^3$Department of Mathematics, Imperial College London\\
{\tt\small \{hchanth, fcendra\}@connect.hku.hk, rubyma@tcl.com, guosheng.yin@imperial.ac.uk, lqyu@hku.hk}
% \and
% \\
% TCL Corporate Research Hong Kong\\
% Hong Kong, China\\
% {\tt\small rubyma@tcl.com}
% \and
% Guosheng Yin\\
% Department of Mathematics\\
% Imperial College London\\
% {\tt\small gyin@hku.hk}
% {\tt\small }
% \and
% Lequan Yu\\
% Department of Statistics and Actuarial Science\\
% The University of Hong Kong\\
% {\tt\small lqyu@hku.hk}
}

\maketitle

\def\thefootnote{*}\footnotetext{The ﬁrst two authors contributed equally to this work.}\def\thefootnote{\arabic{footnote}}

%%%%%%%%% ABSTRACT
\begin{abstract}
Graph-based methods have been extensively applied to whole slide histopathology image (WSI) analysis due to the advantage of modeling the spatial relationships among different entities.
However, most of the existing methods focus on modeling WSIs with homogeneous graphs (\eg, with homogeneous node type).
Despite their successes, these works are incapable of mining the complex structural relations between biological entities (\eg, the diverse interaction among different cell types) in the WSI. 
We propose a novel heterogeneous graph-based framework to leverage the inter-relationships among different types of nuclei for WSI analysis.
Specifically, we formulate the WSI as a heterogeneous graph with ``nucleus-type” attribute to each node and a semantic similarity attribute to each edge.
We then present a new heterogeneous-graph edge attribute transformer (HEAT) to take advantage of the edge and node heterogeneity during massage aggregating.
%the structural information of WSI
%--- a novel architecture designed for graph-based histopathology image analysis. 
%
Further, we design a new pseudo-label-based semantic-consistent pooling mechanism to obtain graph-level features, which can mitigate the over-parameterization issue of conventional cluster-based pooling.
%is based on pseudo-labels generated from a pretrained network and
%
Additionally, observing the limitations of existing association-based localization methods, we propose a causal-driven approach attributing the contribution of each node to improve the interpretability of our framework.
% , which is shown to have improvement over conventional graph explainers based on associations. 
%
Extensive experiments on three public TCGA benchmark datasets demonstrate that our framework outperforms the state-of-the-art methods with considerable margins on various tasks. 
Our codes are available at \href{https://github.com/HKU-MedAI/WSI-HGNN}{https://github.com/HKU-MedAI/WSI-HGNN}.
\end{abstract}
%%%%%%%%% BODY TEXT
%---------------------------------------------------------
\section{Introduction}
%---------------------------------------------------------
% Examination of histopathology slides is extremely 

Histopathology slides provide rich information on
diagnosis and treatment planning for many cancer diseases. 
The recent technological advancements in tissue digital scanners facilitate the development of whole slide histopathology image (WSI) analysis. 
However, traversing through the WSI with diverse magnifications is time-consuming and tedious for pathologists due to the large-scale nature of the WSI (e.g., its typical size is 60,000 $\times$ 60,000 pixels). 
Hence deep learning techniques play an important role as they introduce accurate and automated analysis of WSIs, which can significantly relieve the workload of pathologists.

\begin{figure}
    \centering
    \includegraphics[width=0.5\textwidth]{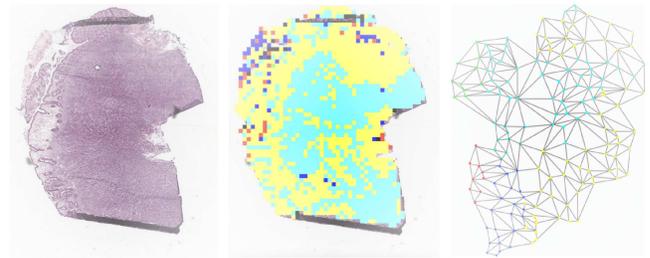}
    \caption{\textbf{Left: } Input WSI. \textbf{Middle: }A WSI with selected patches and associated node types. (Black - no label; cyan - neoplastic; red - inflammatory; blue - connective; yellow - dead; green - non-neoplastic epithelial). \textbf{Right: }Constructed heterogeneous graph with different types of nodes and edge attributes (Illustrative).}
    \label{fig: wsi_with_types}
    %\vspace{-0.2cm}
\end{figure}

Since it is difficult to fit the complete WSI into the memory, most of the works adopt multiple instance learning (MIL) to divide the WSI into instances and then aggregate them for WSI analysis. 
However, these methods operate on bags of instances that do not emphasize the inter-relationships between these instances. 
Recently, the emergence of graph neural networks (GNNs) has made large progress in representing the spatial relationships between instances. 
As a result, there are many attempts to represent the WSIs as graphs of instances. 
Figure \ref{fig: wsi_with_types} presents an example of a graph constructed from WSI.
%learn WSI representation on the compressed graph domain which overcomes the space limitations that CNNs suffer. 
%
%Additionally, GNNs show great improvement in modeling image data over the state-of-the-art (SOTA) convolutional neural networks (CNNs). 
%
Unlike convolutional neural networks (CNNs) that aggregate features based on locality in the Euclidean space, GNNs focus on locality on graph topology, which offers more flexibility in analyzing the deep connections between features in the image data beyond the spatial locality~\cite{ahmedt2021GraphMIAsurvey}. 
For example, GNNs are able to learn relational information and distinguish cells based on their apposition to tumor cells, or normal stroma (i.e., cells which are tumor-infiltrating lymphocytes or from an adjacency inflammatory response), which are important for prognosis~\cite{chen2021patchGCN, saltz2018TILcells}. 

However, existing paradigms on graph-based WSI analysis focus on representing the WSI with a homogeneous graph structure and then predicting the response via vanilla GNNs with cluster-based pooling (i.e., based on similarities of node embeddings). 
%
%These methods pool hierarchical information based on node clusters formed by intermediate node features. They further conduct localization by applying SOTA graph explainers to outline the important regions associated with the prediction.
%
Despite their successes, these methods suffer from several drawbacks: 
(i) GNNs on homogeneous graphs focus on aggregating direct relational information from neighboring nodes, where the complex relational information of the graphs is often neglected.
(ii) For different graphs, the clusters defined by similarities between node embeddings have inconsistent meanings. This introduces a large degree of freedom in parameters and leads to over-parameterization issue~\cite{balaji2020overparameterization}. 
Therefore, GNNs tend to easily overfit due to a lack of identifiability~\cite{gu2021bayesianpyramid}.
% The clusters defined by similarities between node embeddings are inconsistent among different graphs, limiting the identifiability and robustness of the model. 
%(iii) The SOTA graph explainers \cite{luo2020PGExplainers, ying2019gnnExplainer} focus on the association between the nodes and  predictions, where the causal relationship (e.g., which nodes are the cause of the disease) cannot be incorporated by these explainers. 

In view of these limitations, we propose a novel framework for WSI analysis, which leverages a heterogeneous graph to learn the inter-relationships among different types of nodes and edges. 
The heterogeneous graph introduces a ``nucleus-type" attribute to each node, which can serve as an effective data structure for modeling the structural interactions among the nuclei in the WSI. 
To tackle the aggregation process in the heterogeneous graph, we propose a novel heterogeneous-graph edge attribute transformer (HEAT) architecture which can take advantage of the edge and node heterogeneity. 
Thus, the diverse structural relations among different biological entities in the WSI can be incorporated to guide the GNN for more accurate prediction.
% We further design a novel heterogeneous GNN architecture based on transformers which can effectively mine the neighboring and hierarchical information in the graph. 
% Our method is able to mine the meta-relations in the heterogeneous graph by the proposed heterogeneous edge attribute transformer (HEAT) architecture. 
%
Further, to obtain the graph-level representations for slide-level prediction, we propose a semantic-consistent pooling mechanism --- pseudo-label (PL) pooling, which pools node features to graph level based on clusters with a fixed definition (i.e., nucleus type).
The proposed PL pooling can regularize the graph pooling process by distilling the context knowledge (i.e., pathological knowledge) from a pretrained model to alleviate the over-parameterization issue~\cite{balaji2020overparameterization}. 
Additionally, we propose a Granger causality~\cite{granger1969GrangerCausality1} based localization method to identify the potential regions of interest with clinical relevance to provide more insights to pathologists and promote the clinical usability of our approach.
%feasibility of applying our method to clinical scenarios.
% the clinically usability of our approach
%Such a causal localization method produces less noise and is able to outline the region of interest more accurately. The explanations generated by our model provide more insights to pathologists for the nucleus in a WSI that can be potentially causal in disease prediction tasks. which accelerate the progress in finding the causes of diseases, especially for cancer.

We extensively evaluate our method on three TCGA public benchmark datasets, including colon adenocarcinoma cancer (COAD) and breast invasive carcinoma (BRCA) datasets from the TCGA project \cite{weinstein2013TCGA} and the Camelyon 16 dataset \cite{bejnordi2017camelyon16}, and compare to various latest state-of-the-art (SOTA) methods.
% We extensively evaluate our method on the colon adenocarcinoma cancer (COAD) and breast invasive carcinoma (BRCA) datasets from the TCGA project \cite{weinstein2013TCGA}, and the Camelyon 16 dataset \cite{bejnordi2017camelyon16}.
Our method outperforms the competitors on cancer staging, cancer classification, cancer typing, and localization tasks.

% Our code can be found anonymously at \url{https://anonymous.4open.science/r/wsi-het-graph-28E0}.

% Our contributions are summarized as follows:

% \begin{itemize}
%     \item We propose a novel framework to model WSI with a heterogeneous graph, which learns the structural interactions between patches of nuclei.  
%     \item We design a novel heterogeneous GNN architecture based on transformers which can effectively mine the neighboring and hierarchical information in the heterogeneous graph constructed.
%     \item We propose a novel graph pooling method based on node types predicted by a pretrained nuclei classifier model, which demonstrates to be more robust and consistent pooling performance over state-of-the-arts pooling methods. 
%     \item We develop a causality-aware graph explainer to attribute the causal factors related to the classification which can potentially highlight the causal regions of the disease.
%     \item Extensive experiments have been performed on a variety of tasks on different real datasets with comparisons to an array of competitors. Our method outperforms the state-of-the-arts by a significant margin. 
% \end{itemize}

\begin{figure*}[tbp]
    \centering
    \includegraphics[width=1.0\textwidth]{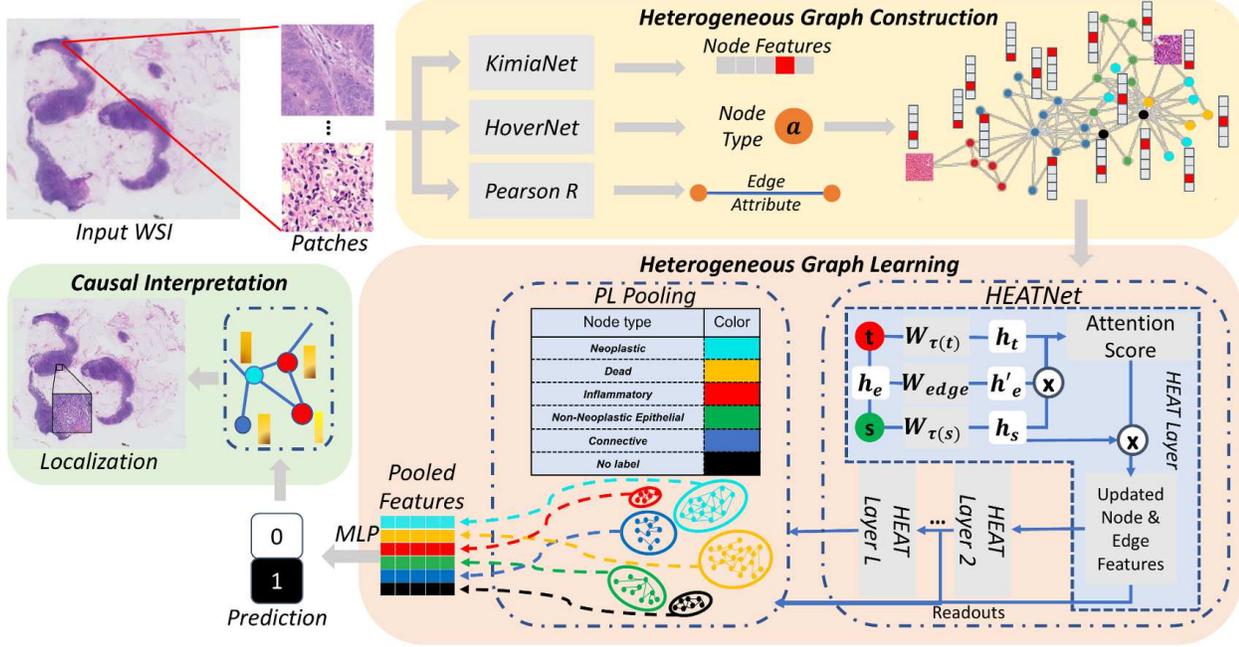}
    \caption{The paradigm of our proposed heterogeneous graph-based WSI analysis framework, which includes heterogeneous graph construction, heterogeneous-graph edge attribute transformer (HEAT) for structural information aggregation, pseudo-label-based (PL) graph pooling for slide-level prediction and casual-driven localization.}
    \label{fig:framework}
\end{figure*}

%---------------------------------------------------------
\section{Related Works}
%---------------------------------------------------------

%--------------------------------
\para{Multiple Instance Learning on WSIs.}
Existing WSI analysis approaches generally adopt MIL
\cite{farris2021MILCNN, tellez2019MILCNN, graham2019hovernet, riasatian2020kimianet, chen2021patchGCN, wang2019rmdl, zheng2022GTNMIL}, which first divide the WSI into fixed-size patches and then compress the information of these patches into low-dimensional vectors. 
Conventional methods aggregate bags of instances to learn WSI-level features for final predictions. 
Tellez \etal \cite{tellez2019MILCNN} compress the WSI-level image into embedding vectors and use a standard CNN to perform patch-level and WSI-level cancer classification. 
These CNN-based methods analyze local areas in the Euclidean space on fixed connectivity (i.e., fixed-size kernels), limiting the performance beyond the spatial locality. 
Graph-based methods \cite{zheng2022GTNMIL, chen2021patchGCN, hou2022h2MIL} have recently been proposed, which model the interactions between instances via graphs. 
Their capability of modeling instances based on graph topology provides more flexibility to analyze complex structures of WSIs. 
Chen \etal \cite{chen2021patchGCN} propose patch-GCN, a method of modeling WSI with homogeneous graphs, and regress survival data with a graph convolutional neural network (GCN) \cite{kipf2016gcn}. 
Zheng \etal \cite{zheng2022GTNMIL} propose a graph-based MIL method using graph transformer networks \cite{yun2019GTN}.  
In spite of their power, most of these WSI methods use homogeneous graphs, which limits the information mined from WSIs. 
A recent method~\cite{hou2022h2MIL} is proposed to model WSIs with heterogeneous graphs, where the heterogeneity in each patch is introduced by different resolution levels. 
However, it only considers the resolution level heterogeneity of patches, with insufficient ability to model the complex contextual interaction between patches in the same resolution level.
%shrinking its capability to model the complex relational information in the WSI.
%
%In this work, we propose a novel method based on heterogeneous graphs, which is shown to outperform SOTA homogeneous-graph-based WSI analysis methods by numerical studies. 

%--------------------------------
\para{Graph Neural Networks.}
Although the SOTA GNNs have shown great successes in many problem domains \cite{kojima2020kgcn, hu2020HGT, lerer2019PBG}, they are mostly focused on homogeneous graphs \cite{kipf2016gcn, velivckovic2017GAT, zhuang2018dualgcn, xu2018GIN, yun2019GTN}. 
These architectures extract the locality information on the graph topology and learn the graph representations by performing aggregation on neighboring nodes. 
However, the potential heterogeneity in nodes and edges is not incorporated by these homogeneous GNN algorithms, and therefore their capability in mining the structural information is limited. 
Several works attempt to address the heterogeneity in their architectural designs  \cite{schlichtkrull2018RGCN, hu2020HGT, wang2019HAN} and assume that the relation type is finite and discrete. 
However, when modeling images with graphs, the heterogeneity in relations is typically continuous (e.g., the similarity between nodes) or high-dimensional. Although there are several attempts \cite{chen2021patchGCN, gong2019EGNN} to extend SOTA GNNs \cite{kipf2016gcn, velivckovic2017GAT} to incorporate edge attributes, their works are limited to homogeneous graphs.  

% There are several works on neural network architectures on heterogeneous graphs. For example, Schlichtkrull \etal propose the relational graph neural network (RGCN) to model knowledge graphs with an edge type linear projection layer to incorporate the heterogeneity in edge types.  

% \citeauthor{wang2019HAN} \cite{wang2019HAN} proposed heterogeneous graph attention network --- an attention-based algorithm to hierarchically extract the node and semantic features in a heterogeneous graph. 

%--------------------------------
\para{Graph Pooling.}
Graph pooling aims to aggregate node-level features to obtain graph-level features. Conventional methods \cite{kipf2016gcn} directly take the average of node-level features to extract graph-level features, which tends to over-smooth the signals of the nodes and cannot generate representative graph-level features. 
Recently, there is extensive development of graph pooling algorithms based on the clusters of the embeddings \cite{eliasof2020diffpool, ranjan2020asap, hou2022h2MIL}. 
However, the clusters constructed based on similarity are inconsistent across graphs. 
This leads to a large degree of freedom in parameters which easily causes overfitting. %
A semantic-consistent pooling method is therefore needed.
%regularize the cluster-based pooling process.
% The adaptive structure aware pooling(ASAP)  algorithm proposed by Ranjan \etal \cite{ranjan2020asap} first learns clusters in an unsupervised method according to the spatial distance of the embeddings in each layer, then the embeddings of each cluster are pooled by a readout layer (e.g., mean or sum of the embeddings). 
% However, these pooling methods heavily rely on the clusters defined on each individual graph which do not have a consistent meaning across graphs. Such limitation severely reduces the identifiability of the trained models and the inconsistency in cluster definition would lead to  poor model performance. Additionally, the aforementioned methods forming clusters are ad-hoc and not related to the problem domain. It is difficult to reason out the mechanism of the pooling, limiting the interpretability of the methods. Observing the limitations, we develop a novel unsupervised method by predefined clusters, which is intrinsically explainable and provide consistent pooling results in graph-level prediction tasks. 
%--------------------------------

\para{Explaining GNNs.}
Despite the success of graph neural networks, their poor interpretability of the parameters makes them notoriously recognized as ``blackboxes". 
With the advances in network attribution methods \cite{schwab2019cxplain}, extensive attempts have been made to open such ``blackboxes" \cite{ying2019gnnExplainer, luo2020PGExplainers}. Generating network explanation is an important qualitative step in the WSI analysis since it can highlight the abnormal regions for further investigation. 
Conventional explainers try to find the associations between the parameters in deep neural networks (or the nodes in GNNs) and the predictions. 
GNNExplainer \cite{ying2019gnnExplainer} is the SOTA method explaining the contributions of node features to the GNN predictions. 
It trains feature masks on each node and edge feature to minimize the prediction loss of a trained GNN. 
PGExplainer \cite{luo2020PGExplainers} shares the same objective as GNNExplainer and trains a generative model to generate explanations.
% The resulting feature mask will be the contribution of each node and edge. 
% PGExplainer \cite{luo2020PGExplainers} borrows the objective from GNNExplainer and makes use of node features in the graph to train a generative neural network to generate explanations for the graph.
%
Recently, there has been emerging attention in generating causal explanations for GNNs \cite{schwab2019cxplain, lin2021gem}, and most of the methods focus on the Granger causality as the explanation objective. 
Gem \cite{lin2021gem} trains explanation generators from the causal perspective. Causal explainers attempt to provide explanations of features that are causal rather than associated with the neural network prediction. 
%Gem \cite{lin2021gem} borrows the idea from PGExplainer to train a generative neural network to generate causal explanations for graphs, with the objective to be the causal contribution of each edge or node in the graph.
% Based on these advances in graph explanation, we design a causal interpretation mechanism to segment the causal regions in the WSI.
\vspace{-0.4cm}
%---------------------------------------------------------
\section{Preliminaries}
%---------------------------------------------------------
\noindent
\textbf{Heterogeneous Graph}: A heterogeneous graph is defined by a graph $\mathcal{G}$ = $(\mathcal{V}, \mathcal{E}, \mathcal{A}, \mathcal{R})$, where $\mathcal{V}, \mathcal{E}, \mathcal{A}$ represent the set of entities (vertices or nodes), relations (edges), and entity types, respectively. 
And $\mathcal{R}$ represents the space of edge attributes. 
For $ v \in \mathcal{V}$, $v$ is mapped to an entity type by a function $\tau(v) \in \mathcal{A}$. An edge $e = (s, r, t) \in \mathcal{E}$ links the source node $s$ and the target node $t$, and $r$ is mapped to an edge attribute by a function $ \phi(e) = r \in \mathcal{R}$. 
Every node $v$ has a $d$-dimensional node feature $x \in \mathcal{X}$, where $\mathcal{X}$ is the embedding space of node features.

% \noindent
% \textbf{Granger Causality} \cite{granger1969GrangerCausality1, lin2021gem}: $x_i$ Granger-causes $y_i$ if we are better able to predict $y_i$ using all available information than the information excluding $x_i$.

\noindent
\textbf{Granger Causality} \cite{granger1969GrangerCausality1, lin2021gem}: Let $\mathcal{I}$ be all the available information and $\mathcal{I}_{-X}$ be the information excluding variable $X$. If we can make a better prediction of $Y$ using $\mathcal{I}$ than using $\mathcal{I}_{-X}$, we conclude that $X$ Granger-causes $Y$.

\noindent
\textbf{WSI Classification}:
Given a WSI $X$ and a heterogeneous graph $\mathcal{G}$ constructed from $X$, we wish to predict the label $y$ with a GNN model $\mathcal{M}$. We also aim to assign an importance score $f(v)$ to each node $v \in \mathcal{V}$ in $\mathcal{G}$ as the causal contribution of each patch to the prediction for localization. 
\vspace{-0.5cm}
%---------------------------------------------------------
\section{Methodology}
%---------------------------------------------------------

%--------------------------------
\subsection{Heterogeneous Graph Construction}
%--------------------------------

\begin{figure}
    \centering
    \includegraphics[width=0.50\textwidth]{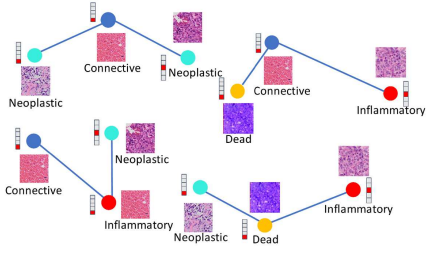}
    \caption{Examples of introduced meta-relations in a heterogeneous graph constructed from a WSI.}
    \label{fig: meta-relation}
    \vspace{-2mm}
\end{figure}

We introduce our methodology of modeling the WSI with a heterogeneous graph. 
Figure \ref{fig:framework} presents the overall workflow of our proposed framework. 
We adopt the commonly used OTSU thresholding algorithm~\cite{chen2021patchGCN} and sliding window strategy to crop each WSI into non-overlapping patches. 
Uninformative patches with backgrounds are removed.
These patches define the nodes of the graph constructed. 
To define the corresponding node type, we use HoverNet \cite{graham2019hovernet} pretrained on the PanNuke dataset \cite{gamper2019pannuke} to classify the patches into predefined types. 
HoverNet detects nuclei in each patch and assigns types to these nuclei. 
By majority votes, we take the most frequently predicted nucleus type to be the type of the patch. 
Figure \ref{fig: wsi_with_types} presents an example of a WSI with patches selected from the OTSU and node types generated by HoverNet \cite{graham2019hovernet}. 
We use a pretrained feature encoder (i.e., KimiaNet \cite{riasatian2020kimianet}) to obtain the embeddings of each patch, which serves as the features of each node in the heterogeneous graph. 

Based on the nodes and node features, we define the edges and edge attributes between the patches. For each node $v \in \mathcal{V}$, we use the $k$-nearest neighbor algorithm to find $k$ nodes that have the most similar features to that node, and connect edges between node $v$ and these neighboring nodes. 
%(i.e., geometrically equivalent to the cosine similarity)
For each edge, we compute the Pearson R correlation between the head and tail node features as the edge attributes. The edge attributes introduce heterogeneity in edges and highlight meta-relations in the WSI. 
We adopt data augmentations (e.g., randomly removing some edges) during training  to alleviate the potential noises introduced by the edge attributes.
%\hchan{Although edge attributes may introduce additional noises to data distribution, can alleviate the noises and mitigate the issue.}
As a result, we obtain a heterogeneous graph $\mathcal{G}$ with heterogeneity introduced by different node types and edge attributes.
As shown in Figure \ref{fig: meta-relation}, a heterogeneous graph outlines the meta-relations between the nuclei in a WSI.
Mining these meta-relations can reveal the structural interactions between the cells, leading to improved performances on different tasks. 
%
% Explicitly specifying the edge attributes between the patches can help the GNN mine the similarity.
%   Learning these meta-relations can compress the structural interactions between nodes into the graph representation. 

% \begin{table}[h]
% \centering
% \caption{HoverNet nuclei types}
% \begin{tabular}{|cc|}
% \hline
% \multicolumn{1}{|c|}{\textbf{Nuclei type}} & \multicolumn{1}{c|}{\textbf{Color}}  \\ \hline
% \multicolumn{1}{|c|}{\textbf{No label}} & \multicolumn{1}{c|}{Black} \\ \hline
% \multicolumn{1}{|c|}{\textbf{Neoplastic}} & \multicolumn{1}{c|}{Cyan} \\ \hline
% \multicolumn{1}{|c|}{\textbf{Inflammatory}} & \multicolumn{1}{c|}{Red} \\ \hline
% \multicolumn{1}{|c|}{\textbf{Connective}} & \multicolumn{1}{c|}{Blue}\\ \hline
% \multicolumn{1}{|c|}{\textbf{Dead}} & \multicolumn{1}{c|}{Yellow}\\ \hline
% \multicolumn{1}{|c|}{\textbf{Non-Neoplastic Epithelial}} & \multicolumn{1}{c|}{Green}\\ \hline
% \end{tabular}\\
% \label{tab: node-classification}
% \end{table}

%--------------------------------
\subsection{Heterogeneous Edge Attribute Transformer}
%--------------------------------
% The definition of $\mathcal{G}$ introduces similarity between the patch embeddings as the attribute of each edge. 
% Edge attributes explicitly specify a prior similarity between nodes, which regularizes the model by introducing prior knowledge.
% In the context of histopathology image analysis, the data are scarce. Hence such a regularization method reduces the model variance and prevents overfitting. 
% The edge attributes also introduce heterogeneity in edges which enables the GNN to mine the structural information in the graph. Hence, simply neglect the edge attributes would lead to information loss and suboptimal model performance. 

The conventional graph attention mechanism is incapable of tackling the heterogeneity of the graph.
Inspired by the transformer architecture \cite{vaswani2017attAlluNeed} and its extension on graphs \cite{yun2019GTN, hu2020HGT, huang2020dahgt}, we propose a new graph aggregation layer, named the Heterogeneous Edge Attribute Transformer (HEAT) layer, to aggregate the structural relations between biological entities in the built heterogeneous graph. 
We explicitly incorporate the node types and continuous edge features into the aggregation process, which guides the learning of edge similarities. 
Our proposed architecture also generalizes the existing architecture to incorporate continuous or high-dimensional edge attributes and simplifies the use of linear layers to avoid overfitting led by model over-parameterizations.
%
%Our proposed model architecture simplifies the existing architecture \cite{hu2020HGT} by preserving the transformation layers only. Such simplification avoids over-specification in model parameters and prevents overfitting. In the context of WSI analysis when data are scarce, avoiding over-specification is vital to model performance.  Improvements in our model over SOTA GNNs based on transformers are shown by numerical studies. 
% We will present in the following section our novel aggregation algorithm incorporating edge attributes, and demonstrate the improvement in performance by numerical studies. 
% We denote the output of the ($l$)-th layer as $H^{(l)}$, which is also the input of ($l+1$)-th layer.
% Information propagation is performed on each edge. 

For each edge $e = (s, r, t)$ and each attention head $i$, we project the target node $t$ into a query vector with a linear projection layer $\bm W^i_{\tau(s)}$, and the source node into a key vector with $\bm W^i_{\tau(t)}$. We also compute the value vector $\bm h_{\text{value}}^i$ of each source node by the same projection layer $\bm W^i_{\tau(s)}$
% We use multi-head attention to compute the interaction between the key and query vectors. Then attention score is computed by the dot product similarity between the key and query vectors, 
% \begin{align*}
%     H^{(l+1)} = \sigma (\Tilde{A} \cdot H^{(l)} \cdot W^{(l)}) \quad 
%     K^i(s) & = \text{K-Linear}^i_{\tau(s)} \Big( H^{(l-1)}[s]\Big) \\
%     Q^i(t) & = \text{Q-Linear}^i_{\tau(t)} \Big( H^{(l-1)}[t]\Big),
% \end{align*}
\begin{align*}
    \bm h_{\text{key}}^i = \bm W^i_{\tau(s)}  H^{(l-1)}_s,  \quad \bm h_{\text{query}}^i & = \bm W^i_{\tau(t)} H^{(l-1)}_t,\\
    \bm h_{\text{value}}^i = \bm W^i_{\tau(s)}  H^{(l-1)}_s, 
\end{align*}
where $H^{(l-1)}_v$ is the input node feature for node $v \in \mathcal{V}$ from the $(l-1)$-th layer. 
These projection layers can project node features of various node types into a node-type-invariant embedding space. 
The edge attributes from the $(l-1)$-th layer $h_e^{(l-1)}$ are also projected to $h'_e = W_{\text{\text{edge}}} h_e^{(l-1)}$ by a linear projection layer $ W_{\text{edge}}$. 
After projecting the node embeddings, we compute the dot-product similarity between the query and key vectors and further multiply the linear transformed edge attribute to the similarity score to incorporate the edge attributes in $\mathcal{G}$. 
We then concatenate the scores from each head and take the softmax of the score (i.e., overweights of incoming edges for all neighboring nodes) to obtain the final attention scores to the value vector $\bm h_{\text{value}}^i$,
\begin{align*}
   \text{Attention}(e) &= \underset{\forall s \in N(t)}{\text{softmax}}\Big( \underset{i \in [1,h]}{\Vert}  \text{ATT}(e, i)\Big),\\
    \text{ATT}(e, i) & = \Big(\bm h_{\text{key}}^i h'_e \bm h_{\text{query}}^i \Big) / \sqrt{d},
\end{align*}
where $N(t)$ is the set of all the source nodes pointing to target node $t$, $d$ is the dimension of node embeddings, $\text{ATT}(e, i)$ represents the $i$-head attention score of edge $e$, $\Vert_{i \in [1,h]}$ is the concatenation operator concatenating the attention scores from all heads and  $\text{Attention}(e)$ represents the final attention score of the edges aggregating all the heads. 
We multiply the attention score obtained by the value vector to obtain the output features. 
By doing so, the output features contain both the node-type and edge-attribute-specific information. Hence the HEAT layer can capture the structural information in $\mathcal{G}$ by transforming the node features from different node types. It can also model different semantic relations since edge attributes are included in the aggregation. 

Finally, we perform target-specific aggregation to update the feature of each target node by averaging its neighboring node features. We concatenate all $h$ attention heads to obtain the attention vector for each pair of source and target nodes. For each target node $t$, we conduct a softmax operation on all the attention vectors from its neighboring nodes and then aggregate the information of all neighboring source nodes of
$t$ together.
The updated node features $H^{(l)}_t$ for $\mathcal{G}_l$ can be represented as  
% \begin{figure}
%     \centering
%     \includegraphics[width=0.51\textwidth]{General/NTPool.pdf}
%     \caption{Mechanism of Pseudo-label Pool}
%     \label{fig: PL-POOL}
% \end{figure}
\begin{align*}
    H^{(l)}_t = \underset{\forall s \in N(t)}{\bigoplus} \Big (\underset{i \in [1,h]}{\Vert} \bm h^i_{\text{value}} \cdot \text{Attention}(e) \Big),
\end{align*}
where $\oplus$ is an aggregation operator (e.g., mean aggregation). 
%The information of all neighboring source nodes of $t$ is then aggregated to $t$, 
The updated graph $\mathcal{G}_l$ is returned as the output of the $l$-th HEAT layer. Algorithm \ref{alg:heat} demonstrates the overall process of our proposed HEAT layer. 
\vspace{-0.3cm}
%\vspace*{-\baselineskip}
\begin{center}
%\scalebox{0.9}{
%\begin{minipage}{\linewidth}
\begin{algorithm}[t]
\begin{algorithmic}[1]
\Statex \textbf{Input:} 
\Statex Heterogeneous graph $\mathcal{G}_{l-1}$ with node features $\{H_i^{(l-1)}, \forall i \in \mathcal{V}\}$ and edge attribute $\{h_e^{(l-1)}, \forall e \in \mathcal{E}\}$; 
\Statex Node-type specific projection layers $\{\bm W^i_{a}, \forall a \in \mathcal{A}\}$
\Statex Edge attribute transformation layer $W_{\text{edge}}$.
\Statex \textbf{Output:} The updated graph $\mathcal{G}_{l}$ with node features $\{H_i^{(l)}, \forall i \in \mathcal{V}\}$, and the edge features  $\{h_e^{(l)}, \forall e \in \mathcal{E}\}$
\State Initialize projection layers for each node type
% \State Initialize scaling parameter $\alpha$
\For{$e = (s,t) \in \mathcal{E}$}
    \State $\bm h_{\text{key}}^i  = \bm W^i_{\tau(s)}  H^{(l-1)}_s$  \Comment{Project the source node}
    \State $\bm h_{\text{value}}^i  = \bm W^i_{\tau(s)}  H^{(l-1)}_s$  \Comment{Compute value vector} 
    \State $\bm h_{\text{query}}^i  = \bm W^i_{\tau(t)} H^{(l-1)}_t$
    \Comment{Project the target node} 
    \State $h'_e \gets W_{\text{edge}} \cdot h_e^{(l-1)}$ 
    \Comment{Project the edge attribute}
    % \State $\bm h_t' \gets \bm h_{\text{key}}^i  \cdot \bm h_{\text{query}}^i \cdot  h'_e$  \Comment{Attention Score}
    \State $\text{ATT}(e, i)  = \Big(\bm h_{\text{key}}^i h'_e \bm h_{\text{query}}^i \Big) / \sqrt{d}$  
    \State $\text{Attention}(e) = \underset{\forall s \in N(t)}{\text{softmax}}(\Vert_{i \in [1,h]}  \text{ATT}(e, i))$
    \State $h_e^{(l)} \gets h'_e$ \Comment{Compute latent edge features}
\EndFor
\For{$t \in \mathcal{V}$}
    \State $H^{(l)}_t = \oplus_{\forall s \in N(t)} ({\Vert}_{i \in [1,h]} \bm h^i_{\text{value}} \cdot \text{Attention}(e))$ 
\EndFor
\State \Return $\mathcal{G}_{l}$
\end{algorithmic}
\caption{The HEAT algorithm.}
\label{alg:heat}
\end{algorithm}
%\end{minipage}%
%}
\end{center}
\vspace{-0.15cm}
%--------------------------------
\subsection{Pseudo-label Graph Pooling}
%--------------------------------

We introduce a novel pooling method --- pseudo-label (PL) pooling, to aggregate information with respect to the pseudo-labels (i.e., node types) predicted from a pretrained teacher network (e.g., HoverNet \cite{graham2019hovernet}).  
Unlike conventional methods of pooling features based on clusters, we define clusters using a pretrained node classifier. 
Pooling from pseudo-labels ensures the semantic consistency in cluster definitions and distills the context knowledge (e.g., nuclei features) from the teacher network.
Specifically, for each node type $a$, we pool all node features belonging to type $a$ into a single vector $\bm h_a$ with a readout layer. 
The pooled features from each node type are then aggregated into a feature matrix $\bm S \in \mathbb{R}^{|\mathcal{A}| \times d}$. 
The graph level feature is then determined by another readout layer (e.g., mean readout).

% Pooling by node types can distill context knowledge (e.g., nuclei types) into the pooling layers. 
Algorithm \ref{alg:PLPool} presents the workflow of the proposed PL Pooling. 
By pooling with the pseudo-labels, we are able to cluster patch representation according to nuclei types, such that the graph-level features are enhanced with the prior knowledge on nuclei type distributions. 
The detailed mechanism of the PL Pool is presented in the supplementary materials. 
We also perform an ablation study in Table \ref{tab: ablation Pooling} and show that PL Pooling outperforms existing pooling methods in cancer classification tasks. 
%\begin{center}
%    \scalebox{0.9}{
%\begin{minipage}{\linewidth}
\begin{algorithm}[t]
\begin{algorithmic}[1]
\Statex \textbf{Input:} Heterogeneous graph $\mathcal{G}$ with node features $\{H_i, \forall i \in \mathcal{V}\}$ and node type set $\mathcal{A}$.
\Statex \textbf{Output:} The pooled graph-level feature $\bm S \in \mathbb{R}^{|\mathcal{A}| \times d}$.
\State Initialize readout layers for each node type $a \in \mathcal{A}$.
\State Initialize aggregate feature matrix $\bm S$.
\For{$a \in \mathcal{A}$}
    \State $X_a \gets $ feature matrix of nodes of type $a$ 
    \State $\bm h_a \gets$ readout$_a$($X_a$)  \Comment{Pool feature with readout layer}
    \State $\bm S_a = \bm h_a$ \Comment{Assign pooled feature to the $a$-th row of $\bm S$}
\EndFor
\State \Return $\bm S$
\end{algorithmic}
\caption{The PL-Pool Algorithm}
\label{alg:PLPool}
\end{algorithm}
%\vspace{-0.45cm}
%    \end{minipage}%
%    }
%\end{center}

\begin{table*}[t]
    \centering
        \begin{tabular}{clccc||cccc}
        \toprule
        \multicolumn{2}{c}{}
        & \multicolumn{3}{c}{\textbf{Cancer Staging (Four Stages)}}
        & \multicolumn{3}{c}{\textbf{Cancer Classification}} 
        \\
        \hline
        \multicolumn{1}{l}{} &
        \multicolumn{1}{l}{\textbf{Model}} &
        \multicolumn{1}{c}{\textbf{AUC}} & 
        \multicolumn{1}{c}{\textbf{Accuracy}} &
        \multicolumn{1}{c}{\textbf{Macro-F1}} &
        \multicolumn{1}{c}{\textbf{AUC}} & 
        \multicolumn{1}{c}{\textbf{Accuracy}} & 
        \multicolumn{1}{c}{\textbf{Macro-F1}}
        \\ \hline
        &ABMIL \cite{ilse2018abmil} 
        & 53.8 (3.7) & 19.2 (7.8)  & 35.8 (4.4) 
        & 97.7 (2.3) & 98.3 (0.9)  & 95.8 (2.2) \\
        &DSMIL \cite{li2021dsmil} 
        & 59.3 (1.4) & 35.7 (5.7) & 37.9 (2.8) 
        & 99.7 (0.2) & 98.6 (0.5) & 96.9 (0.9) \\
        &ReMix \cite{yang2022remix} 
        & 58.3 (1.5) & 33.9 (7.8) & 24.8 (7.5)
        & 94.3 (3.4) & 96.0 (4.6) & 92.8 (5.9) \\
        &PatchGCN \cite{chen2021patchGCN} 
        & 62.5 (4.9) & 38.2 (3.1) & 38.5 (5.7)
        & 91.1 (5.3) & 97.1 (2.0) & 98.8 (1.0) \\
        & GTNMIL \cite{zheng2022GTNMIL} 
        & 54.2 (2.6) & 29.3 (1.4) & 24.3 (3.9)
        & 97.3 (2.6) & 98.1 (1.3) & 95.9 (2.4) \\
        \rot{\rlap{\textbf{TCGA--COAD}}}
        &H$^2$-MIL \cite{hou2022h2MIL} 
        & 58.6 (2.7) & 38.5 (5.4) & 33.0 (5.0)
        & 99.7 (0.4) & 99.2 (0.5) & 97.4 (1.7) \\%\hline
        &\multicolumn{1}{c}{\textbf{HEAT (Ours)}} 
        & \textbf{63.4 (2.5)} & \textbf{40.0 (2.1)}   & \textbf{41.3 (2.7)}
        & \textbf{99.9 (0.2)} & \textbf{99.9 (0.3)} & \textbf{99.2 (0.4)}  \\ \hline
        &ABMIL \cite{ilse2018abmil} 
        & 54.7 (4.6) & 19.0 (10.0) & 23.9 (3.2)
        & 97.3 (1.7) & 98.3 (1.1) & 97.3 (1.6) 
        \\
        &DSMIL \cite{li2021dsmil} 
        & 51.4 (4.7) & 18.3 (14.9) & 23.2 (2.3)  
        & 98.7 (0.5) & 95.6 (1.4) & 93.3 (2.0) 
         \\ 
        &ReMix \cite{yang2022remix} 
        & 58.8 (2.2) & 35.6 (16.2) & 27.6 (5.8) 
        & 96.1 (0.7) & 95.8 (2.6) & 93.0 (3.4)  
        \\
        &PatchGCN \cite{chen2021patchGCN} 
        & 50.3 (0.2) & 41.6 (0.5) & 25.1 (0.3)
        & 96.2 (1.7) & 98.2 (0.8) & 98.4 (0.8) 
        \\
        &GTNMIL \cite{zheng2022GTNMIL} 
        & 53.0 (3.7) & 41.3 (4.4) & 25.1 (2.3) 
        & 94.7 (1.0) & 94.5 (0.2) & 93.7 (1.7) 
        \\ 
        \rot{\rlap{\textbf{TCGA--BRCA}}}
        &H$^2$-MIL \cite{hou2022h2MIL} 
        & 52.1 (7.2) & 53.7 (2.6) & 21.2 (2.5) 
        & 97.9 (2.7) & 98.0 (1.5) & 97.6 (2.2)  
        \\%\hline
        &\multicolumn{1}{c}{\textbf{HEAT (ours)}}
        & \textbf{61.9 (3.8)} & \textbf{55.8 (6.4)} & \textbf{27.7 (16.3)}
        & \textbf{98.8 (0.7)} & \textbf{98.3 (0.5)} & \textbf{99.5 (0.7)}
        \\ %\hline
        \bottomrule
        \end{tabular}
    \caption{Cancer staging and classification results [\%] of various methods on TCGA--COAD and TCGA--BRCA datasets.}
    \label{tab:results_quantitative_COAD_BRCA}
    %\vspace{-1.3mm}
\end{table*}

%--------------------------------
\subsection{Prior Knowledge Regularization}
%--------------------------------
Here we discuss the motivation for introducing prior knowledge in our proposed HEAT and PL pooling algorithms. 
In the context of WSI analysis when the data are scarce, while data distributions are sparse and high-dimensional. The curse of high dimensionality makes the sampling distributions difficult to approximate the properties of true distributions of the WSIs. This leads to a significant gap between training and testing distributions.
Hence regularization techniques are needed to reduce the model variance and mitigate performance deterioration when transferring the model from training to testing environments.
Since WSI data contain enriched prior knowledge (e.g., the interaction among different cell types), integrating such knowledge into the framework regularizes the model, such that the testing performance improves. 
Therefore, we design the above two designs by integrating prior knowledge into the feature aggregation procedure. 
Specifically, for the HEAT layer, we integrate the prior knowledge of node type and node attributes when extracting node-level features. 
For PL Pooling, we pool node-level features using prior definitions on node clusters. 
Moreover, we perform data augmentations (e.g., random pruning on edges and nodes) to regularize the learning from training distributions.
Besides that, other regularization such as imposing a Gaussian prior on the model weights (i.e., using a Bayesian neural network) would also achieve the goal.

%--------------------------------
\subsection{Causal-driven Localization}
%--------------------------------
We make use of the Granger causality to outline causal regions in the WSI with the causal graph explainer~\cite{lin2021gem}. 
Given a trained GNN model $\mathcal{M}$, the causal contribution of each node $v$ is given by 
\begin{equation} \label{eq:causal explain}
    \Delta_{\delta, v} = \mathcal{L}(y, \Tilde{y}_{\mathcal{G}}) - \mathcal{L}(y, \Tilde{y}_{\mathcal{G} \text{\textbackslash} \{ v\}}),
\end{equation}
where $y$ is the true label and $\Tilde{y}_{\mathcal{G}} = \mathcal{M}(\mathcal{G})$ and $\Tilde{y}_{\mathcal{G} \text{\textbackslash} \{ v\}} = \mathcal{M}(\mathcal{G} \text{\textbackslash} \{ v\})$ are the predicted labels from $\mathcal{M}$ with input graphs $\mathcal{G}$ and $\mathcal{G} \text{\textbackslash}\{ v\}$, respectively. 
The causality heatmap of the patches can then be visualized with the causal contribution computed for each patch (i.e., node). Addressing causality in instance interpretation can adjust for observational and selection biases, which would improve the explanation accuracy. 
Moreover, the causal property of the explainer could facilitate pathologists to find out potential biomarkers for diagnosis and prognosis by highlighting the patches with clinical relevance in the WSI.

% If $\Delta_{\delta, v} > 0$, the GNN model is performing worse when the node is excluded, indicating that node $v$ is causally contributed to the prediction. 

\begin{table}[h]
    \centering
        \begin{tabular}{lccc p{1.3cm}}
        \toprule
        \multicolumn{1}{l}{\textbf{Model}} &
        \multicolumn{1}{c}{\textbf{AUC}} & 
        \multicolumn{1}{c}{\textbf{Accuracy}} &
        \multicolumn{1}{c}{\textbf{Macro-F1}} 
        \\ \hline
            ABMIL \cite{ilse2018abmil} & 79.5 (7.5) & 80.3 (8.4) & 81.3 (7.4)  \\
            DSMIL \cite{li2021dsmil} & 92.5 (1.7) & 87.3 (2.0) & 86.3 (2.0) \\
            ReMix \cite{yang2022remix} & 92.5 (7.2) & 90.0 (8.1) &  90.3 (7.7) \\
            PatchGCN \cite{chen2021patchGCN}  & 88.6 (3.5)  & 92.1 (2.3) &  92.3 (2.4) \\
            GTNMIL \cite{yun2019GTN} & 89.7 (4.7) & 81.2 (4.8) & 89.2 (4.9) \\
            H$^2$-MIL \cite{hou2022h2MIL}  & 92.1 (3.9)  & 88.2 (5.8)  & 88.0 (5.8) \\ \hline
            \textbf{HEAT (ours)}  & \textbf{92.8 (2.5)} & \textbf{92.7 (2.2)} & \textbf{93.3 (1.9)} \\
            % \hline
        \bottomrule
        \end{tabular}
     \caption{Cancer typing results [\%] of our method compared to various methods on the TCGA--ESCA dataset.}
    \label{tab: results_typing_ESCA}
    %\vspace{-4mm}
\end{table}
%---------------------------------------------------------
\section{Experiments}
%---------------------------------------------------------

%--------------------------------
\subsection{Datasets}
%--------------------------------

We use WSIs from the public TCGA--COAD (cancer staging task: $1304$ cases, classification task: $1434$ cases), TCGA--BRCA (cancer staging task: $1328$ cases, classification task: $1712$ cases), and TCGA--ESCA (typing task: $213$ cases) from the TCGA project \cite{weinstein2013TCGA} and Camelyon 16 \cite{bejnordi2017camelyon16} as the benchmark datasets.
On average, around 300 patches are sampled from each WSI in the TCGA datasets (around 5,000 for Camelyon 16), where each patch corresponds to a node in the final heterogeneous graph.
For the TCGA--COAD and the TCGA--BRCA datasets, we conduct two tasks for the benchmark methods --- cancer staging and cancer classification. 
For the cancer staging task, all the cases are divided into the ``Stage I", ``Stage II", ``Stage III", and ``Stage IV" classes. 
For the cancer classification task, all the cases are divided into the ``Normal" and ``Tumor" classes. 
For the cancer typing task, we use TCGA--ESCA dataset where all the cases are divided into two classes i.e., ``Type I: adenocarcinoma" and ``Type II: squamous cell carcinoma".   % For the Camelyon 16 dataset, we design a cancer classification task tottt rk our method to its competitors. 
We also evaluate the localization ability of our framework on the Camelyon 16 dataset, as this dataset provides the tumor mask annotations.                     %
A detailed summary of datasets is provided in supplementary materials. 

\subsection{Implementation Details}
%--------------------------------

The proposed framework is implemented in Python with the \textit{Pytorch} library on a server equipped with four NVIDIA TESLA V100 GPUs. 
We use \textit{openslide} \cite{goode2013openslide} as the tool to process the WSIs.  
The dropout ratio of each dropout layer is selected as 0.2. 
All models are trained with 150 epochs with early stopping. 
The batch size is selected as 2. 
%The testing model checkpoint is selected from the checkpoint having the highest validation area under the receiver operating characteristic curve (AUC). 
We adopt the cross-entropy loss to train the network for classification tasks. 
We use the Adam optimizer to optimize the model with a learning rate of $5 \times 10^{-5}$ and a weight decay of $1 \times 10^{-5}$. 
We perform data augmentations on the training graphs by randomly dropping the edges and nodes, and adding Gaussian noises to the node and edge features.
%
%More network architecture details and hyperparamters can be found in supplementary materials.

\begin{table}[t]
    \centering
        \begin{tabular}{lccc p{1.3cm}}
        \toprule
        \multicolumn{1}{l}{\textbf{GNN Architecture}} &
        \multicolumn{1}{c}{\textbf{AUC}} &
        \multicolumn{1}{c}{\textbf{Accuracy}} &
        \multicolumn{1}{c}{\textbf{Macro-F1}}
        \\ \hline
            GCN \cite{kipf2016gcn} & 90.8  & 90.9 & 90.0 \\
            GAT \cite{velivckovic2017GAT} & 85.8 & 86.4 & 88.9 \\
            GIN \cite{xu2018GIN} & 91.6 & 90.9 &  83.3 \\
            HetRGCN \cite{schlichtkrull2018RGCN}  & 82.5 & 83.3 &  88.9 \\
            HGT \cite{hu2020HGT} & 87.8 & 87.5  & 83.3 \\ \hline
            \textbf{HEAT (ours)} & \textbf{92.8 } & \textbf{92.7} & \textbf{93.2} \\
            % \hline
        \bottomrule
        \end{tabular}
    \caption{Cancer typing results [\%] of our method compared to various GNN architectures on the TCGA--ESCA dataset.}
    \label{tab: ablation GNNs}
    % \vspace{-1mm}
\end{table}

\begin{table}[t]
    \resizebox{\linewidth}{!}{\def\arraystretch{1.05}
    \centering
        \begin{tabular}{lccc p{1.3cm}}
        \toprule
        \multicolumn{1}{l}{\textbf{Pooling Method}} &
        \multicolumn{1}{c}{\textbf{AUC}} & 
        \multicolumn{1}{c}{\textbf{Accuracy}} &
        \multicolumn{1}{c}{\textbf{Macro-F1}} 
        \\ \hline
        Sum pooling & 95.5 & 99.3 & 99.2 \\
        Max pooling & 95.1 & 98.6 & 99.2 \\
        Mean pooling & 97.7 & 95.8 & 99.8 \\
        Global attention pooling \cite{li2015globalAttentionPooling} & 94.7 & 97.9 & 99.2 \\
        IH-Pool \cite{hou2022h2MIL} & 99.3 & 97.2 & 88.1 \\
        ASAP \cite{ranjan2020asap} & 99.2 & 98.6 & 95.1 \\ \hline
        % DiffPool \cite{eliasof2020diffpool} & &    \\ \hline
        \textbf{PL-Pool (ours)} & \textbf{99.6} & \textbf{99.3} & \textbf{99.8} \\ 
        % \hline 
        \bottomrule
        \end{tabular}}
    \caption{Cancer classification results [\%] on TCGA--COAD of our pooling method to various comparable pooling methods using GCN and KimiaNet feature encoder.}
    \label{tab: ablation Pooling}
    %\vspace{-1mm}
\end{table}

%--------------------------------
\subsection{Experiment Settings and Evaluation Metrics}
%--------------------------------
We compare our method with an array of SOTA methods, including MIL or graph-based methods. We use five-fold cross-validation to evaluate the overall performance of our framework and other methods. We used the pretrained KimiaNet as the feature extraction for all methods for a fair comparison.
The details of compared methods are listed below.
\begin{itemize}[noitemsep,topsep=0.3pt]
    \item ABMIL \cite{ilse2018abmil}: a MIL framework aggregating bag-level instance information by the attention mechanism. 
    \item DSMIL \cite{li2021dsmil}: a dual-stream multiple instance learning method using max pooling and attention to aggregate the signals from the individual patches.
    \item ReMix \cite{yang2022remix}: a general and efficient MIL's based framework for WSI analysis that takes the advantage of data augmentation and reduces method to produce rich features. 
    \item PatchGCN \cite{chen2021patchGCN}: a hierarchical graph-based model on survival data with patient-level and WSI-level aggregations. We adapt this method as a GCN model with global attention pooling \cite{li2015globalAttentionPooling}. 
    \item GTNMIL \cite{zheng2022GTNMIL}: a graph-based MIL method based on the  graph transformer network \cite{yun2019GTN}.
    \item H$^2$-MIL \cite{hou2022h2MIL}: a tree-graph-based multiple instance learning method that utilizes different magnification levels to represent hierarchical features. 
\end{itemize}

For the cancer staging, classification and typing tasks, we use AUC, classification accuracy, and macro F-1 score as the evaluation metrics. 
Percentage [\%] values are reported for each of the metrics. 
Standard errors are reported in brackets. 
For all metrics, a higher value indicates a better performance. 
Detailed definitions of the evaluation metrics can be found in the supplementary materials.

\begin{table}[t]
    \centering
        \begin{tabular}{lccc p{1.3cm}}
        \toprule
        \multicolumn{1}{l}{\textbf{Balanced dataset}} &
        \multicolumn{1}{c}{\textbf{AUC}} &
        \multicolumn{1}{c}{\textbf{Accuracy}} &
        \multicolumn{1}{c}{\textbf{Macro-F1}}
        \\ \hline
            TCGA--COAD & 99.1 (1.8)  & 99.1 (1.8) & 99.2 (1.7) \\
            TGCA--BRCA & 98.7 (2.5) & 98.7 (2.5) & 98.7 (2.6) \\
            % \hline
        \bottomrule
        \end{tabular}
    \caption{Cancer classification results [\%] of our method on TCGA--COAD and TCGA--BRCA balanced datasets.}
    \label{tab: ablation data imbalance}
    %\vspace{-2mm}
\end{table}

%--------------------------------
\subsection{Comparison with Other Methods}
%--------------------------------

\para{Quantitative Results.}
Table~\ref{tab:results_quantitative_COAD_BRCA} shows the cancer staging and classification results on the TCGA--COAD and the TCGA--BRCA datasets, and Table \ref{tab: results_typing_ESCA} presents cancer typing results on the TCGA--ESCA dataset. 
Compared to graph-based WSI analysis methods \cite{zheng2022GTNMIL, hou2022h2MIL, chen2021patchGCN}, our method demonstrates improved performance, which indicates our graph modeling method potentially better represents the interaction of patches in a WSI than existing graph-based methods.
% We test our method against the competitor having the closest performance to ours~\ylq{need to specify}. The highest p-value we obtain is $0.0328$ when comparing our method to all competitors on all tasks, which is statistical significant under the significance level of $0.05$. The detailed test procedure are shown in the supplementary materials.
%
We also observe that aggregation on a graph of instances is more effective than aggregation on bags of instances in the staging tasks, which implies graph-based methods are more capable of capturing the global information of WSI for staging tasks than conventional MIL methods \cite{li2021dsmil, ilse2018abmil, yang2022remix}. 
We further compare HEAT on the BRCA subtyping task with a recent SOTA method on WSI --- hierarchical image pyramid transformer (HIPT)~\cite{chen2022HIPT}.
Our method achieves an AUC of 89.69 (SD: 3.63), which outperforms the AUC of 87.4 (SD: 6.0) by HIPT.

Additionally, we perform a t-test on the AUCs to demonstrate the statistical significance of our improvements over the SOTA methods, for which the results are presented in Table \ref{tab: t-test}. 
We observe that the improvements are statistically significant over most of the baseline methods under the 0.05 significance level.

\begin{table}[!h]
    \centering
        \vspace{-0.5mm}
            \scalebox{0.9}{
        \begin{tabular}{lcccc p{1.3cm}}
        \toprule
        \multicolumn{1}{l}{\textbf{Methods}} &
        \multicolumn{1}{c}{\textbf{COAD-S}} &
        \multicolumn{1}{c}{\textbf{COAD-C}} &
        \multicolumn{1}{c}{\textbf{BRCA-S}} & 
        \multicolumn{1}{c}{\textbf{BRCA-C}}
        \\ \hline
        ABMIL & 1.91e-16  & 0  & 5.75e-5 & 6.08e-6 \\
        DSMIL & 0.0005  &  0.0333 & 3.26e-9 & 0.3790\\
        ReMix &  1.36e-5 & 0  & 0.0759 & 1.2e-16 \\
        PatchGCN &  0.2899 &  0 & 5.14e-11 & 2.74e-15 \\
        GTNMIL &  2.7e-15 &  0 & 5.58e-7 & 2.94e-35\\
        H$^2$-MIL & 4.5e-5  & 0.0343  & 3.94e-8 & 0.0082\\
        % ABMIL & 0  &  0 & 0 & 0\\
        % DSMIL & 0.0005  &  0.0332 & 0 & 0.3790\\
        % ReMix &  0 &  0 & 0.0477 & 0\\
        % PatchGCN &  0.2899 &  0 & 0 & 0\\
        % GTNMIL &  0 & 0  & 0 & 0\\
        % H$^2$-MIL & 0  &  0.0343 & 0 & 0.0082\\
            % \hline
        \bottomrule
        \end{tabular}}
    \caption{P-values of two-sample t-tests on AUCs between our method and baselines (S: cancer staging; C: cancer classification).}
    %\vspace{-1mm}
     \label{tab: t-test}
\end{table}

% \begin{figure*}[t]
%     \centering
%     \includegraphics[width=0.75\textwidth]{Results/explanations/WSI_explainer_comparison.pdf}
%     \caption{The input WSI (left) and the explanation heatmaps generated by GNNExplainer (middle) and our causal-driven explanation method (right). Ground truth regions are outlined with red boundaries. Lighter yellow indicates a higher importance score. }
%     \label{fig:gnn_explainer}
% \end{figure*}

\para{Qualitative Results.} 
We compute the causal contribution of each patch using Equ.~\eqref{eq:causal explain}.  
We visualize the patch image associated with that node to outline the causal regions related to the predictions. 
We also compare our causal explanation method to numerous baseline graph interpretation methods based on associations \cite{ying2019gnnExplainer}. 
Figures in the supplementary materials present the explanation results with different graph explainers on the Camelyon 16 dataset.
It is observed that using an association-based explainer provides a smooth heatmap where many regions are highlighted as important. 
A such heatmap is less accurate in localizing the tumor regions and pathologists still need to traverse a large number of abnormal regions suggested by the explainer to identify tumor regions. 
On the contrary, we observe that using a causal explainer can outline the tumor regions in the WSIs more accurately, with the heatmap more concentrated on the ground-truth tumor regions compared to association-based explainers (e.g., GNNExplainer~\cite{ying2019gnnExplainer}). 

\subsection{Analysis of Our Framework}
%---------------------------------------------------------
%We perform analysis to our framework to discuss potential key changes to different components of our framework, and how these changes affect the model performance. 

\para{Effectiveness of Heterogeneous Graph Construction.}
We compare our method with other SOTA GNNs \cite{hu2020HGT, kipf2016gcn, velivckovic2017GAT, schlichtkrull2018RGCN, xu2018GIN} to evaluate the effectiveness of our heterogeneous graph construction. 
For heterogeneous graph transformer (HGT)~\cite{hu2020HGT} and HetRGCN~\cite{schlichtkrull2018RGCN}, we define the discrete edge types --- each relation either has the ``positive" type representing positive correlations between the nodes of the edge, or the ``negative" type representing negative correlations. 
Table~\ref{tab: ablation GNNs} presents cancer typing results of our method compared to various SOTA GNN aggregation methods on the TCGA--ESCA dataset. 
Not only our method outperforms SOTA homogeneous GNN architectures~\cite{kipf2016gcn, xu2018GIN, velivckovic2017GAT}, but it is also superior to some recently heterogeneous GNN architectures~\cite{hu2020HGT, schlichtkrull2018RGCN}.  
This implies the advantage of our proposed architecture for graph-based WSI analysis. 

% We also observe that heterogeneous GNN architectures (HGT and HEAT) outperforms homogeneous GNN architectures (GAT, GCN, and GIN) in general. This demonstrates that modelling a WSI with a heterogeneous graph can better represent the WSI. The heterogeneity highlights structural interaction between different regions of WSI and thus the model learns more representative features.  

\para{Analysis of Different Pooling Strategies.}
We compare our proposed pooling strategy to a variety of comparable pooling methods, including basic pooling methods, such as sum/max/mean poolings and advanced pooling strategies~\cite{hou2022h2MIL,ranjan2020asap}. 
%
%Most of these methods pool node features by local clusters. 
%
Table~\ref{tab: ablation Pooling} presents the comparison results of cancer classification on TCGA--COAD dataset.
We fix the model architecture to be GCN \cite{kipf2016gcn} and the feature encoder as KimiaNet \cite{riasatian2020kimianet}. 
It is observed that our pooling strategy outperforms the competitors, which validates the advantage of using semantic-consistently defined clusters in pooling. 

\para{Performance on Different Class Distributions.}
We observe the WSI datasets for cancer classification is imbalanced (i.e., approximately ten cancer WSIs to one normal WSI). We thus compose a balanced dataset (i.e., normal:cancer = 1:1) with the undersampling strategy to study how the difference in class distributions affect the performance of our model. 
Table~\ref{tab: ablation data imbalance} presents the comparison. It is observed that our method achieves similar performance with the unbalanced setting (See Table~\ref{tab:results_quantitative_COAD_BRCA}).

\para{Generalizability.}
The pretrained features are a key component of our proposed framework.
As the pretrained embedding models are from a diverse WSI context, they can extract good features from most of the WSI datasets.
Because the PanNuke dataset~\cite{gamper2020pannuke} (used to pretrain the HoverNet node type classifier) contains WSIs of most of the common cancer types, this leads to a broad generalization of HoverNet.
%
%The pretrained HoverNet can thus generalize well.
%
Furthermore, one may adopt contrastive learning to fine-tune the pretrained models to improve their generalizability to new datasets in potential deployment scenarios. 

\para{Accuracy of HoverNet.}
The performance of the HoverNet classifier would influence the sensitivity of our framework. 
Since the PanNuke dataset contains WSIs of most of the common cancer types and cohorts of the TCGA dataset (e.g., COAD), there are domain overlaps between them. 
Hence the HoverNet trained on the PanNuke dataset can be transferred to the TCGA dataset for patch types classification with good performance.
%
% Moreover, data augmentation such as node feature dropout can adjust the bias so that our method is still robust when the node type prediction is unsatisfactory.
%
%
% Furthermore, since current clustering methods deliver poor performance as they are unsupervised and based on embedding similarities, using a HoverNet classifier can easily outperform the accuracy of existing methods. 
Furthermore, we perform cancer classification on COAD using node types generated by unsupervised K-means clustering.
The performance (AUC: 98.5) is lower than that using HoverNet predicted node types (AUC: 99.9).
This demonstrates that incorporating the pretrained HoverNet outperforms unsupervised methods and improves WSI analysis.

%---------------------------------------------------------
\section{Conclusion}
%---------------------------------------------------------
We present a novel heterogeneous graph-based framework for WSI analysis.
By modeling WSI as a heterogeneous graph with various node types and edge attributes, our method not only leverages the locality information, but also mines the complex relational information of WSI.
We further design a novel heterogeneous edge attribute transformer architecture to aggregate the structural information in the graph and a semantic consistent pooling method to address the potential over-parameterization problems in conventional pooling.
We provide a causal explanation mechanism to highlight the causal contributions of the instances to improve the clinical usability of our work.
%, which can potentially identify the causes of the diseases in a WSI when performing disease classification. 
%
Extensive experiments on public datasets validate the effectiveness of our proposed framework and our framework could be adapted to other graph-based computer vision tasks, such as 3D point cloud analysis and anomaly detection. %\hchan{In the future, it is meaningful to further develop causal localization methods on WSI, and provide insights to discoveries of disease causes.}

\para{Acknowledgement.} We thank the anonymous reviewers and the area chair for their insightful comments on our manuscript. 
This work was partially supported by the Research Grants Council of Hong Kong (17308321), the Theme-based Research Scheme (T45-401/22-N), the National Natural Science Fund (62201483), and the HKU-TCL Joint Research Center for Artificial Intelligence sponsored by TCL Corporate Research (Hong Kong).

%%%%%%%%% REFERENCES

{\small
\bibliographystyle{unsrt}
\bibliography{references}
}
% \newpage
% \input{6-appendx}

\end{document}